\documentclass[10pt,twocolumn,letterpaper]{article}

\usepackage{iccv}
\usepackage{times}
\usepackage{epsfig}
\usepackage{graphicx}
\usepackage{amsmath}
\usepackage{amssymb}
\usepackage{tablefootnote}
\usepackage{multirow}

\usepackage[pagebackref=true,breaklinks=true,letterpaper=true,colorlinks,bookmarks=false]{hyperref}

\iccvfinalcopy 

\def\iccvPaperID{****} 

\ificcvfinal\fi

\begin{document}

\title{Attending Generalizability in Course of  Deep Fake Detection by Exploring Multi-task Learning}

\author{Pranav Balaji\\
BITS Pilani\\
Hyderabad, India\\
{\tt\small f20190040@hyderabad.bits-pilani.ac.in}
\and
Abhijit Das\\
BITS Pilani\\
Hyderabad, India\\
{\tt\small abhijitdas2048@gmail.com}
\and
Srijan Das\\
University of North Carolina at Charlotte\\
North Carolina, United States\\
{\tt\small sdas24@uncc.edu}
\and
Antitza Dantcheva\\
INRIA\\
Biot, France\\
{\tt\small antitza.dantcheva@inria.fr}
}

\maketitle
\ificcvfinal\thispagestyle{empty}\fi


\begin{abstract}

This work explores various ways of exploring multi-task learning (MTL)  techniques aimed at classifying videos as original or manipulated in cross-manipulation scenario to attend generalizability in deep fake scenario. The dataset used in our evaluation is FaceForensics++, which features 1000 original videos manipulated by four different techniques, with a total of 5000 videos. We conduct extensive experiments on multi-task learning and contrastive techniques, which are well studied in literature for their generalization benefits. It can be concluded that the proposed detection model is quite generalized, i.e., accurately detects manipulation methods not encountered during training as compared to the state-of-the-art.

\end{abstract}

\section{Introduction}

Deepfakes are computer-generated videos, images, or audio recordings that have been manipulated using artificial intelligence and machine learning algorithms to create realistic content of a person doing or saying something they did not actually do \cite{das2021demystifying}. Deepfakes use deep learning techniques, such as neural networks, to manipulate existing content and create something new. The term "deepfake'' is a combination of "deep learning" and "fake". Deepfakes are threats to society because they can be used to spread misinformation, manipulate public opinion, harass individuals, and even blackmail people. They can be particularly dangerous when used in political contexts, where they can be used to damage reputations or influence election outcomes \footnote{https://tinyurl.com/mrxtxk5z}.
Deepfakes can be generated in several ways, manipulation methods involve \textit{generative adversarial network}s (GANs) to create realistic videos \cite{rossler2019faceforensics++}. Another method involves using facial detection technology to map a person's face onto a face or to \textit{superimpose} their face onto an existing video. One common approach is to train an autoencoder to reconstruct a specific person’s face on any given image of a body. 

Due to the recent fuel in media tampering techniques such as deep fake, corresponding \textit{manipulation-detection approaches} have been developed. Manipulation detection techniques included image-based \cite{rossler2019faceforensics++}, video-based \cite{das2021demystifying, roy20223d}, or jointly audio and video-based \cite{conti2022deepfake} approaches.  In the context of image-based deep fake detection to explore the angle of generalizability, second-order local anomaly detection has been used \cite{fei2022learning} and self-consistency is explored in the work of \cite{tiwari2022occlusion}. Singular frame based detection techniques ensemble predictions across frames of a video \cite{zhao2021multi, coccomini2022combining}. While computationally efficient, they do not exploit the presence of temporal inconsistencies \cite{mirsky2021creation}. Hence recently, video-based generalized deep fake detection has been gaining significance. This for the detection of temporal inconsistencies with respect to the lip movement \cite{agarwal2020detecting}, jitters between frames \cite{guera2018deepfake} and optical flow \cite{amerini2019deepfake}. Identity consistency has also been explored, with \cite{dong2022protecting} using a transformer to identify inconsistency between the inner and outer face region based on a database of known identities. Authors of \cite{agarwal2019protecting} modelled behaviors of world leaders from recorded stock footage and identify behavioral inconsistencies in deepfakes. These techniques usually require prior identity or behavior information about the victim, so they are suited for celebrities but do not scale to civilian victims.

Tolosana et al. \cite{tolosana2020deepfakes} reviewed manipulation technique such as DeepFake in 2020 w.r.t. facial regions, and fake detection performance and concluded that the generalization of such detection methods is challenging.
In other words, when detection methods, such as the presented ones are confronted with adversarial attacks, outside of the training set, such networks have a dramatic drop in performance. Challenge of generalization of deep fake is studied in \cite{das2021demystifying, roy20223d}. In \cite{chen2022self} self-supervised learning (SSL) has been explored by learning adversarial examples for generalized deep fake detection. Further, in the same direction of research, a multimodal approach has been adopted in \cite{conti2022deepfake, zhou2021joint} using audio-video analysis. The angle of generalizability for deep fake detection is not much explored and yet remain to be a challenge.

Hence, in order to address the unsolved challenge of generalizability in deep fake detection, we explore the concept of MTL encompassing both supervised and self-supervised learning (SSL) approaches~\cite{he2020momentum}. By employing MTL, we aim to not only detect deep fakes but also identify the specific type of deep fake jointly. MTL involves jointly learning multiple tasks, typically with a shared early layer or common connections at the beginning of the network and individual task-specific layers at the end. This shared layers in early processing allows to improved learning by sharing parameters across tasks. Convolutional Neural Networks (CNNs), based on deep neural networks (DNNs), have demonstrated exceptional performance in simultaneously solving diverse tasks within the MTL framework~\cite{misra2016cross}.

MTL has been extensively studied in various domains of machine learning and deep learning applications. For instance, it has found application in natural language processing tasks, such as unified representations~\cite{collobert2008unified} and representation learning~\cite{liu2015representation}. Additionally, MTL has been employed in speech recognition~\cite{deng2013new}, drug discovery~\cite{Ramsundar2015MassivelyMN}, and computer vision-related tasks including face analysis~\cite{das2018mitigating, happy2020apathy}, pedestrian detection~\cite{chen2014joint, yin2017multi, tian2015pedestrian}, face alignment~\cite{zhang2016joint}, attribute prediction~\cite{abdulnabi2015multi}, among others.
Furthermore, MTL has gained significance in recent times for face attribute learning, also known as semantic features, as they provide a more natural description of objects~\cite{niu2019robust} and activities~\cite{happy2019characterizing}. This approach enables a comprehensive understanding of the visual world by jointly modeling multiple attributes in face-related tasks.  
In their recent work \cite{foggia2023multi, das2018mitigating}, the authors propose a biasless approach for face attribute analysis. These couple of work on MTL inspires and gives the importance of achieving generalizability in deep fake detection, it becomes crucial to explore MTL. While MTL is commonly studied using Supervised Learning (SL), the exploration of MTL in the context of SSL remains relatively unexplored, making it highly relevant to the problem at hand. Additionally, it is worth noting that deep fake detection has not been extensively investigated within the MTL framework. Therefore, in this paper, we aim to address the identified research gap by exploring deep fake detection in the MTL scenario. 


SSL has demonstrated successful applications in both image-based methods such as SimCLR \cite{chen2020simple}, MoCo \cite{he2020momentum}, MAE \cite{he2022masked}, DINO \cite{caron2021emerging}, and video-based approaches like CoCLR \cite{han2020self}, VideoMAE \cite{tong2022videomae}, SVT \cite{ranasinghe2022self}. SSL focuses on learning representations by leveraging inherent data structure, eliminating the reliance on labeled data. This paradigm has shown significant improvements in generalizability, with recent studies surpassing the performance of supervised models on zero-shot testing \cite{he2022masked}. Building on these advancements, we anticipate that SSL can enable the model to learn better representations for enhancing generalizability in the context of deep fake detection. Specifically, the SSL-based approach can empower the model to effectively discriminate between fake and real instances, irrespective of the manipulation technique employed.
This advancement raises several open questions regarding the application of MTL and SSL compared to SL in deepfake detection. Thus, in this paper, we aim to address the following questions:
\begin{itemize}
    \item Can MTL be effectively employed for deepfake detection?
    \item How does the choice of loss affect MTL?
    \item How can a combination of Contrastive and Cross entropy-based learning be utilized to enhance the generalizability of deepfake
detection? 
    \item What are the appropriate sub-tasks and their relationships with the primary task within the MTL framework for deepfake detection?
\end{itemize}

To address the aforementioned inquiries, we conducted a thorough examination of the existing literature on MTL, SL, and SSL. In our experimental investigation, we explored both contrastive learning and SL techniques within the MTL framework. The outcomes of our study indicate that MTL holds significant value for the task of deepfake detection.
However, the subsequent question arises: which type of learning, SL or SSL, is most relevant to this particular problem? Our experimental analysis provides insights into this query, revealing that incorporating a combination of SL or/and SSL as a sub-task of manipulation classification yields superior results compared to employing SSL alone for the detection task within the MTL paradigm.
Finally, we also introspect the relation of the sub-task (see Figure 1) and the binary classification problem (fake vs real). 

Therefore, in this study, we comprehensively explore the optimal strategies for employing MTL in the detection of deepfakes. Through extensive experimental analysis, we demonstrate various configurations of MTL networks that yield exceptional performance in deepfake detection. Furthermore, we offer end users a range of options for selecting the backbones of MTL for traditional deepfake detection networks in both Contrastive and Cross entropy approaches (see Figure 2), tailored to their specific constraints.
We firmly believe that this empirical investigation not only contributes to the advancement of deepfake detection and its generalizability but also presents promising avenues for future research in this domain.


\section{Proposed methodology}

As mentioned previously that Deepfakes can be generated in several ways and several new techniques can evolve. Main manipulation methods involve using generative and superimpose technique. Detecting deepfakes is a challenging problem because they are designed to be convincing and difficult to distinguish from real content. Moreover, generalizability of deep fake detection  i.e while they are tested on cross-manipulation technique is an added challenge.  To underpin this problem we hypothesized to use explore MTL in both supervised learning (SL) and self-supervised learning (SSL) as a base while learning detection technique and type of deep fake jointly. In lieu of these we assume that in the course of MTL-based training, the model will learn a better representation for generalizability. In other words, the MTL-base will help the model learn how to discriminate between fake and real irrespective of the manipulation used.

\subsection{MTL for deep fake detection}


The aim of the experiments is to design a training paradigm that enables the core encoder to learn features that are generalizable across various manipulation techniques. To this end, we employ 2 techniques: Multi-task learning and SSL with Momentum contrast (MoCo) with label-informed positive and negative pool construction. Due to popularity and the relevance in the learning aspect MoCo was selected among the SSL techniques available. The following encoder and stream format is employed throughout the paper.

\textbf{Encoder:} In all our experiments, the S3D encoder, pre-trained with CoCLR \cite{han2020self} is used.\\

\textbf{Classifier:} A 3 layer MLP was used to classify the embedding vector produced by the S3D into the respective classes. This was trained using Cross entropy loss.\\

\textbf{Dual stream:} To encourage the encoder to learn not just to differentiate between original videos and manipulated videos but also to identify the type of manipulation, we consider training it using a stream for Multi-class classification. We proceed to explain MTL in details.\\
\begin{figure}[ht]
\centering
\label{fig:multi_task}
\includegraphics[width=\columnwidth]{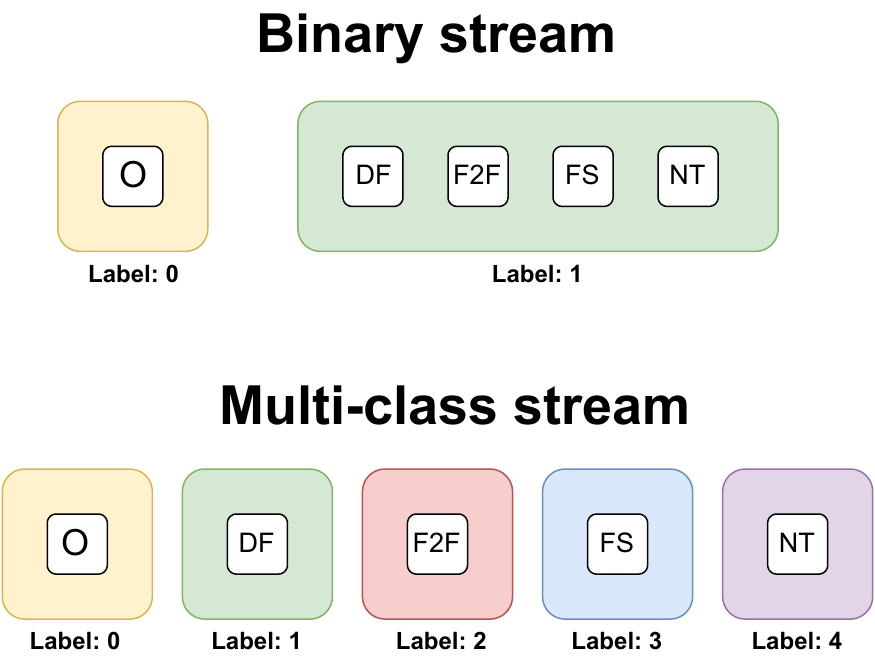}
\caption{Difference between Binary stream and multi-class stream}
\end{figure}
\begin{figure*}
\centering
\label{fig:all_cases}
\includegraphics[width=\textwidth]
{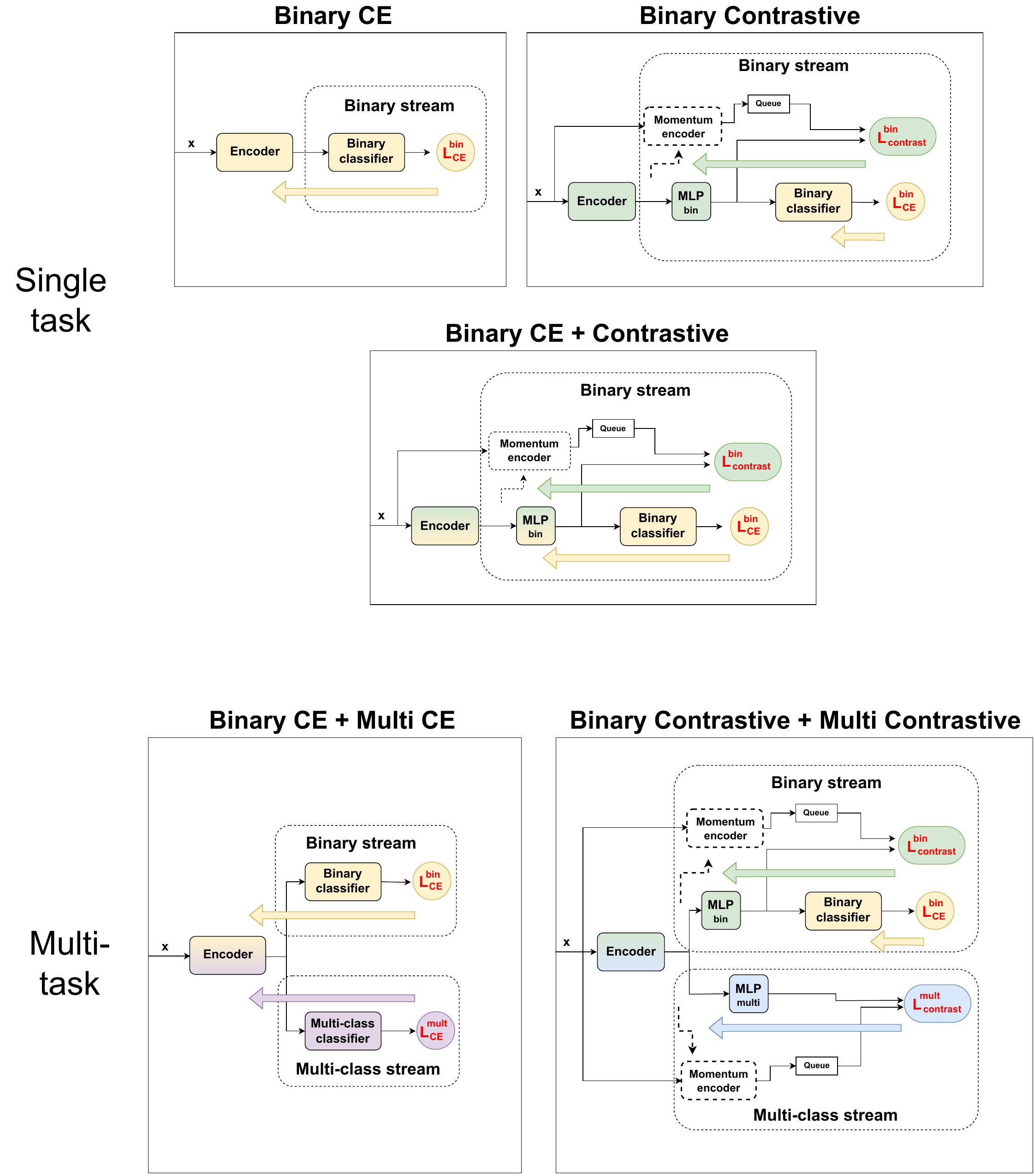}
\caption{Block representation of different scenarios of MTL proposed for deep fake detection. Colored arrows indicate gradient flow}
\end{figure*}

\subsubsection{Multi-task learning}
This involves using two different streams - binary and multi-class, where labels are defined differently.\\
\textbf{Binary stream}: The task performed is binary classification where 0 represents "original" media while 1 represents "manipulated" media. All manipulation techniques - Deepfakes, Face2Face, FaceSwap and NeuralTextures fall under this category. This stream is of ultimate interest.\\
\textbf{Multi-class stream}: The task performed is multi-class classification where 0 represents "originals" and each of classes 1-n are assigned to different manipulation techniques. For example, in the case that a model is being trained on Originals, FaceSwap, Face2Face and NeuralTextures, there would be a total of 4 classes. This stream gives the model richer information in the form of \textit{which} manipulation technique is being encountered, which could help in generalizability. Both types of the models are illustrated in Figure 1. This stream is only employed during training and is discarded during testing.

\subsubsection{Label-Informed MoCo}
Following the success in applying MoCo \cite{he2020momentum} to labelled data \cite{han2020self}, we design the other technique.
In essence, it is MoCo for videos applied while the positive and negative pool is constructed not by augmentations, but by indexing the corresponding labels. The positive pool for a given sample, in this case, may consist of more than one instance as opposed to the original MoCo paradigm. Hence, we used the Multi-Instance Info-NCE loss as per the following equation:
\begin{equation}
\label{eq:nce_loss}
\begin{split}
    L_{contrast} = \\
    -E [ \log \frac{\Sigma_{p \in P_i}\exp(z_i \cdot z_p/\tau)}{\Sigma_{p \in P_i}\exp(z_i \cdot z_p/\tau) + \Sigma_{n \in N_i}\exp(z_i \cdot z_n / \tau)} ]
\end{split}
\end{equation}
Equation \ref{eq:nce_loss}: Multi-Instance Info-NCE loss. The numerator is a sum of 'similarity' between sample $x_i$ and its positive set $P_i$. $P_i$ is defined as the set of examples in the queue with the same label as $x_i$, and $N_i$ is defined to be its complementary set. $z_i$ is the representation of $x_i$ from the main encoder while $z_p$ and $z_n$ are taken from the queue, generated by the EMA encoder in previous iterations.
$z_i \cdot z_j$ refers to the similarity between vectors $z_i$ and $z_j$ and is defined to be the dot product. Intuitively, this loss has the effect of pulling the representations of positive pairs together and pushing the negative pairs apart.

The construction of positive and negative pools for MoCo is also based on the definition of labels which is different in the binary and multi-class streams. In either case, positive samples are drawn from the same pool as the anchor while negative samples are drawn from any random pool that is different from the anchor.

Six different cases were considered for training, with respect to gradient propagation to the main encoder:

\begin{itemize}
    \item \textbf{Binary CE:} The encoder is updated using the Binary Cross Entropy (CE) loss $L^{bin}_{CE}$. This entails standard fine-tuning. The model was pretrained with CoCLR SSL weights.
    \item \textbf{Binary Contrastive:} The encoder is updated only using the binary version of $L_{contrast}$ defined in \ref{eq:nce_loss}, $L^{bin}_{contrast}$, where positive and negative pairs are defined as in Figure 2. Inspired by MoCo v2 \cite{chen2020improved}, we use an MLP head after the encoder wherever $L_{contrast}$ is involved. The performance is monitored using a standalone Binary Classifier updated using BCE loss. These gradients are cut off at the classifier.
    \item \textbf{Binary CE + Contrastive:} Similar to the previous case, but here the gradients from the Binary CE (BCE) loss are not cut off at the classifier and are allowed to propagate to the main encoder. Hence, the encoder is updated using both $L^{bin}_{CE}$ and $L^{bin}_{contrast}$.
    \item \textbf{Binary CE + Multi CE:} With the addition of multi-tasking, an additional stream performing multi-class classification is introduced. The encoder is updated from the Cross Entropy loss from both streams, $L^{bin}_{CE}$ and $L^{mult}_{CE}$. The binary stream is of primary focus at the end of training.
    \item \textbf{Binary Contrastive + Multi Contrastive:} The encoder is updated using the two label-informed contrastive losses from the two streams, $L^{bin}_{contrast}$ and $L^{mult}_{contrast}$ where the difference is between positive and negative pair construction as defined in Figure 2. Just like in the case of \textit{Binary Contrastive}, Separate MLP heads are used in each stream and a standalone Binary Classifier is trained with $L^{bin}_{CE}$ to monitor performance.
\end{itemize}

All block diagrams for all the cases are illustrated in Figure 2.

\begin{figure}[ht]
\centering
\label{fig:ff_examples}

\includegraphics[scale=0.4]{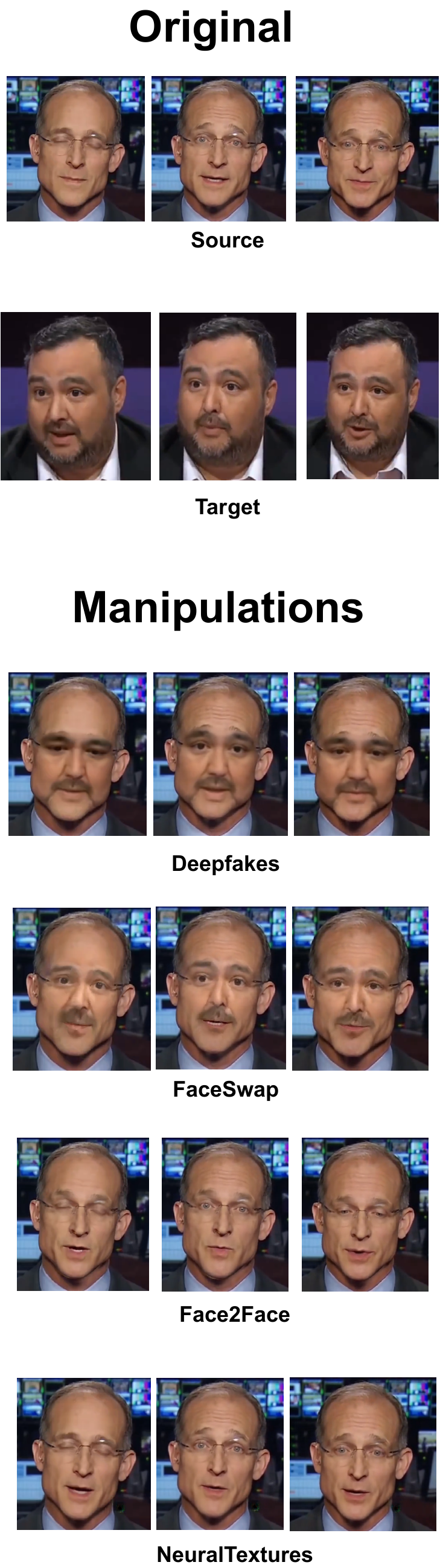}
\caption{Examples of Manipulation techniques in FaceForensics++}
\end{figure}

\begin{table*}
\label{tab:ff++}
\center
\caption{Statistics of different manipulating techniques available in Faceforensics++}

\begin{tabular}{||c | c | c | c | c | c | c ||} 
 \hline
 \textbf{Split} & \textbf{DeepFake} & \textbf{Face2Face} & \textbf{FaceSwap} & \textbf{NeuralTextures} & \textbf{Original} & \textbf{Total} \\ [1.0 ex] 
 \hline
\textbf{Train}         & 720               & 720                & 720               & 720                     & 720               & 3600 \\ 
 \hline
\textbf{Val}           & 140               & 140                & 140               & 140                     & 140               & 700 \\
 \hline
\textbf{Test}          & 140               & 140                & 140               & 140                     & 140               & 700 \\
 \hline
 \textbf{Total}   & 1000              & 1000               & 1000              & 1000                    & 1000              & 5000 \\
 \hline
\end{tabular}%
\end{table*}

\section{Experiment results and discussion}

In this section we proceed to explain the employed database and the experimental results achieved to validate our proposed methodology. 

\subsection{Dataset}
We employed a publicly available and most popular dataset on Deep fake.

\textbf{FaceForensics++}: The FaceForensics++(FF++) dataset \cite{rossler2019faceforensics++} is a large-scale benchmark dataset for face manipulation detection, which was created to help develop automated tools that can detect deepfakes and other forms of facial manipulation. The dataset consists of more than 1,000 high-quality videos with a total of over 500,000 frames, which were generated using various manipulation techniques such as facial reenactment, face swapping, and deepfake generation.

The videos in the dataset are divided into four categories, each corresponding to a different manipulation technique: Deepfakes(DF), Face2Face(F2F), FaceSwap(FS), and NeuralTextures(NT). Deepfakes use machine learning algorithms to generate realistic-looking fake videos, while Face2Face and FaceSwap involve manipulating the facial expressions and identity of a person in a video. NeuralTextures uses a different approach by altering the texture of a face to make it appear different.
The dataset includes both real and manipulated videos, with each manipulation technique applied to multiple individuals.
Examples are shown in Figure 3.

\begin{table*}[t]
\label{tab:results}
\center
\caption{FF++ results for cross manipulation scenario.}
\begin{tabular}{|c|c|c|c|c|}
\hline
\textbf{Train on}               & \textbf{Test on}               & \textbf{Type of training} & \textbf{Accuracy}         \\ \hline
F2F, FS, NT            & DF                    & Baseline(S3D)          & 76.08\%          \\ \hline
F2F, FS, NT            & DF                    & Binary CE          & 80.08\%          \\ \hline
F2F, FS, NT            & DF                    & Binary Contrastive          & 84.77\%          \\ \hline
F2F, FS, NT            & DF                    & Binary CE + Contrastive          & 81.64\%          \\ \hline
F2F, FS, NT            & DF                    & Binary CE + Multi CE          & \textbf{85.16\%}          \\ \hline
F2F, FS, NT            & DF                    & Binary Contrastive + Multi Contrastive          & 83.20\%          \\ \hline
F2F, FS, NT            & DF                    & DFDC Winner \tablefootnote{The same model used in DFDC's winning solution, but without the DFDC pre-training} \cite{seferbekov2020dfdc}      & \textbf{85.71\%}          \\ \hline
                       &                       &                  &                  \\ \hline
DF, FS, NT             & F2F                    & Baseline(S3D)          & 66.08\%          \\ \hline                       
DF, FS, NT             & F2F                   & Binary CE          & \textbf{73.44\%}          \\ \hline
DF, FS, NT             & F2F                   & Binary Contrastive         & 68.36\%          \\ \hline
DF, FS, NT             & F2F                   & Binary CE + Contrastive          & 66.02\%          \\ \hline
DF, FS, NT             & F2F                   & Binary CE + Multi CE          & 68.75\%          \\ \hline
DF, FS, NT             & F2F                   & Binary Contrastive + Multi Contrastive          & 58.98\%          \\ \hline
DF, FS, NT             & F2F                   & DFDC Winner \cite{seferbekov2020dfdc}          & 72.86\%          \\ \hline
                       &                       &                  &                  \\ \hline
DF, F2F, NT            & FS                    & Baseline(S3D)          & 75.12\%          \\ \hline
DF, F2F, NT            & FS                    & Binary CE          & 79.30\%          \\ \hline
DF, F2F, NT            & FS                    & Binary Contrastive          & 82.42\%          \\ \hline
DF, F2F, NT            & FS                    & Binary CE + Contrastive          & \textbf{85.55\%}          \\ \hline
DF, F2F, NT            & FS                    & Binary CE + Multi CE          & 83.59\%          \\ \hline
DF, F2F, NT            & FS                    & Binary Contrastive + Multi Contrastive          & 80.08\%          \\ \hline
DF, F2F, NT            & FS                    & DFDC Winner \cite{seferbekov2020dfdc}          & 51.07\%          \\ \hline
\multicolumn{1}{|l|}{} & \multicolumn{1}{l|}{} &                  &                  \\ \hline
DF, F2F, FS            & NT                    & Baseline(S3D)          & 69.18\%          \\ \hline
DF, F2F, FS            & NT                    & Binary CE          & \textbf{75.78\%}          \\ \hline
DF, F2F, FS            & NT                    & Binary Contrastive          & 69.92\%          \\ \hline
DF, F2F, FS            & NT                    & Binary CE + Contrastive          & 70.31\%          \\ \hline
DF, F2F, FS            & NT                    & Binary CE + Multi CE          & 73.44\%          \\ \hline
DF, F2F, FS            & NT                    & Binary Contrastive + Multi Contrastive          & 64.06\%          \\ \hline
DF, F2F, FS            & NT                    & DFDC Winner \cite{seferbekov2020dfdc}          & 60.00\%          \\ \hline
\end{tabular}
\end{table*}

\begin{table}[t]
\label{tab:results}
\center
\caption{Average of all FF++ results for cross manipulation scenario.}
\begin{tabular}{|c|c|}
\hline
\textbf{Type of training} & \textbf{Avg. Accuracy}         \\ \hline
Baseline(S3D)          & 71.62\%          \\ \hline
Binary CE          & 77.15\%          \\ \hline
Binary Contrastive          & 76.17\%          \\ \hline
Binary CE + Contrastive          & 75.98\%          \\ \hline
Binary CE + Multi CE          & \textbf{77.74\%}          \\ \hline
Binary Contrastive + Multi Contrastive          & 71.58\%          \\ \hline
DFDC Winner \cite{seferbekov2020dfdc}      & 67.41\%          \\ \hline
\end{tabular}
\end{table}

\begin{figure*}
\centering
\label{fig:misclassifications}
\includegraphics[width=\textwidth]{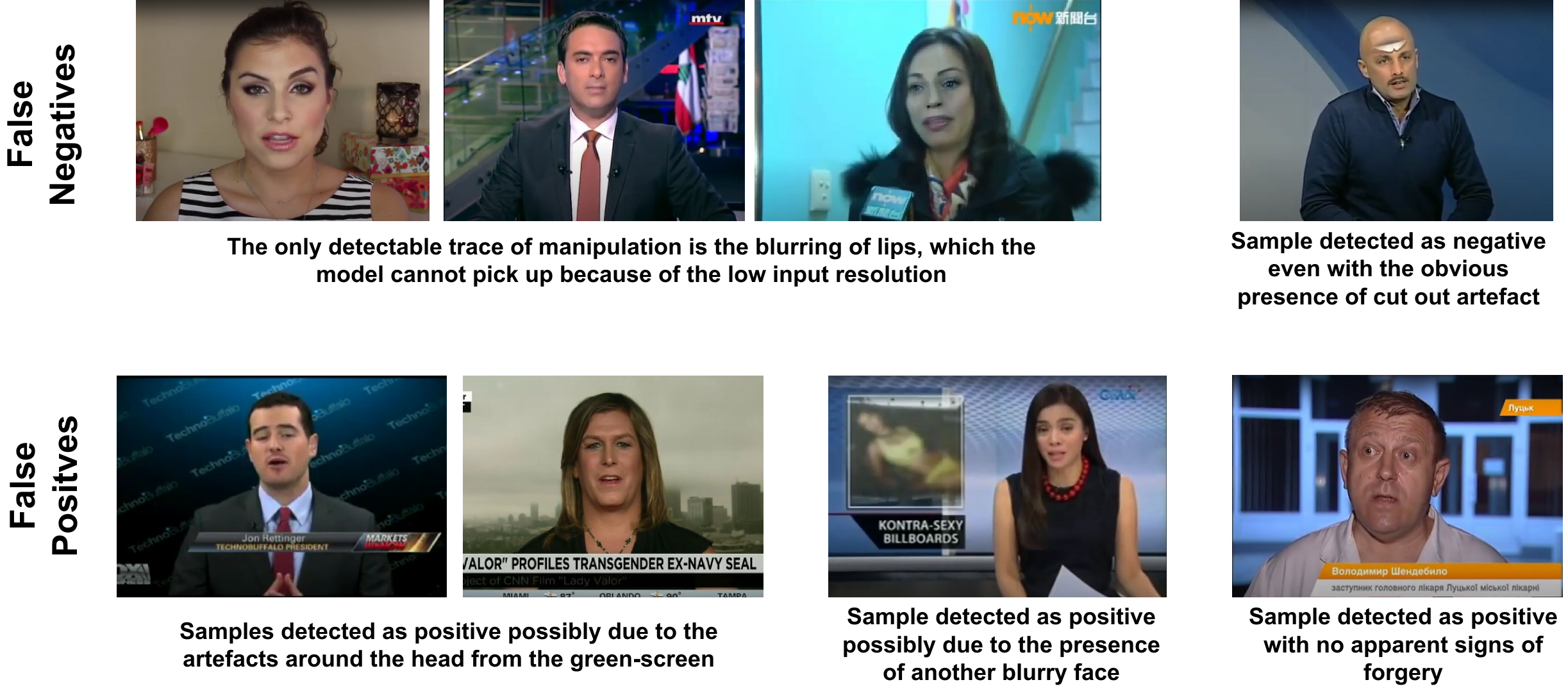}
\caption{Examples of mis-classifications.}
\end{figure*}

\subsection{Discussion}

Evidently, MTL scenarios seem to be outperforming the base line i.e. S3D initialized with Supervised weights and the state-of-the-art for deepfake detection, the winner \cite{seferbekov2020dfdc} of the DFDC Challenge \cite{dolhansky2020deepfake}. \cite{seferbekov2020dfdc} differs from our models in 3 key ways: Architecture (Image based EfficientNet B7 \cite{tan2019efficientnet} vs our video based S3D \cite{xie2018rethinking}), Pre-training (Noisy student \cite{xie2020self} vs our CoCLR \cite{han2020self}) and Image resolution (380 vs our 128).
Our models outperform \cite{seferbekov2020dfdc} in the majority of the scenarios. Only for the \textit{Deepfakes} cross-manipulation scenario,\cite{seferbekov2020dfdc} narrowly outperformed our best model by ~0.5\%.\\
It is interesting to note that all video-based models significantly outperform \cite{seferbekov2020dfdc} on the \textit{cross FaceSwap} and \textit{cross NeuralTextures} cases even though they operate on less than half the frame resolution. This leads to the conclusion that at least for these manipulation techniques, \textbf{\textit{temporal information is much more relevant for identifying forgery than frame resolution.}}\\
Moreover, all methods involving MTL or contrastive learning suffer on the \textit{cross Face2Face} and \textit{cross NeuralTextures} scenario compared to regular BCE training. Analyzing the difference between the various manipulation techniques in Figure 3, it is clear that Face2Face and NeuralTextures work by changing the lip movement to match the target, rather than transplant the target's face on the source video. The general difficulty in detecting these techniques in cross training evaluation suggest that these techniques need special consideration and cannot be generalized to, from the other techniques.\\
As long as MTL is not involved, Contrastive loss leads to better generalization on Deepfakes and FaceSwap, while CE loss leads to better generalization on the other two.

Now we proceed to answer the question raised in the introduction section:

\begin{itemize}
    \item \textbf{Can MTL be effectively employed for deepfake
detection?} MTL helps in certain cases as it can be evident from Table 3 that Binary CE + Multi CE produces the best average results in cross-manipulation scenario. Although, just Binary CE training comes very close.\\
Therefore, \textit{\textbf{Yes in certain cases, but narrowly.}}

\item \textbf{How does the choice of loss affect MTL?}

Supervised learning with Cross Entropy loss performs best, as it can be evident from Table 3. In CE training, Binary CE + Multi CE produces the best average results while Binary CE comes close.\\
However, in the contrastive loss case, MTL significantly degrades the results as shown by Binary Contrastive + Multi Contrastive being significantly worse than Binary Contrastive training.\\
Therefore, \textit{\textbf{Cross Entropy loss is suited for Multi-task learning while Contrastive loss is not.}}

\item \textbf{How can a combination of Contrastive and Cross entropy-based learning be utilized to enhance the generalizability of deepfake
detection?} It is evident from Table 3 that jointly training on CE and Contrastive loss is worse than training with a single type of loss. This leads us to the conclusion that the two losses produce conflicting gradients that lead the model to a sub optimal parameter space.\\
Therefore, \textit{\textbf{It is not good to jointly train on Cross Entropy and Contrastive loss.}}

\item \textbf{What are the appropriate sub-tasks and their relationships with the primary task within the MTL framework
for deepfake detection?} From Table 3, it is evident that identifying the type of manipulation is a good sub-task when using Cross Entropy loss, but not when using Contrastive loss.\\
However, there could be better sub-tasks for generalization such as non-contrastive SSL approaches used as Self-Supervised Auxiliary Training \cite{das2023few}, some examples being SVT \cite{ranasinghe2022self} and VideoMAE \cite{tong2022videomae}, which we leave to future work.
Therefore for now, \textit{\textbf{Finding the type of manipulation is a good sub task for Cross Entropy training.}}\\
\end{itemize}

A few examples of mis-classifications are shown in Figure 4. Analyzing the false negatives show that sometimes, the only detectable trace of manipulation is the blurring of lips, and the models are unable to pick it up because of the low input resolution. A straightforward fix to this would be to use a higher resolution model, at the cost of higher compute.\\
False positives on the other hand happen due to various reasons. Sometimes the green-screen effect is present around the head region, which the model could mistake as artefacts from forgery techniques. This brings into question if such media should really be treated as \textit{original}. If not, a larger training set containing more of such examples could be used. Sometimes, there are other blurred faces in the video which could confuse the model. This could be resolved by robust face detection and alignment, which we leave to future work.

\section{Conclusion}

Overall, this study shows that treating deep fake detection naively as a video classification problem rather than image classification greatly helps in generalizability.
Manipulation detection and classifying the type of manipulation present are proved to be related tasks by the observation that one serves as a good sub-task for the other. Moreover, we show that while both Cross Entropy training and Contrastive training are good, jointly training on the two losses seems to be counterproductive. When it comes to the contrastive loss framework, we have shown that it plays poorly into multi-task learning, at least for our chosen sub-task of identifying the type of manipulation present.  While our proposed methods did not surpass the state-of-the-art on FaceForensics++ for all scenarios, we have shown that our models trained on a subset of forgery techniques can generalize to never-before-seen manipulations much better than the state-of-the-art.

\section{Acknowledgements}

We acknowledge the high performance computing facility, Sharanga, at the Birla Institute of Technology and Science - Pilani, Hyderabad campus primarily used for our experiments. We also acknowledge the support provided by INRIA as part of an internship to Pranav Balaji.

{\small
\bibliographystyle{ieee_fullname}

}

\end{document}